\documentclass{article}

\PassOptionsToPackage{numbers, sort&compress}{natbib}

     \usepackage[preprint]{neurips_2021}

\usepackage[utf8]{inputenc} 
\usepackage[T1]{fontenc}    
\usepackage{hyperref}       
\usepackage{url}            
\usepackage{booktabs}       
\usepackage{amsfonts}       
\usepackage{nicefrac}       
\usepackage{microtype}      

\usepackage{graphicx}
\usepackage{multirow}
\usepackage{amsmath}

\usepackage{subfig}
\usepackage{booktabs}
\usepackage{framed,multirow}
\usepackage{makecell}
\usepackage{placeins}
\usepackage[dvipsnames]{xcolor}
\usepackage{multirow}
\title{Deep Superpixel Generation and Clustering for Weakly Supervised Segmentation of Brain Tumors in MR Images}

\author{
  Jay J.~Yoo\textsuperscript{1,2,4,6},
   Khashayar Namdar\textsuperscript{1,2,6}, 
   Farzad Khalvati \textsuperscript{1,2,3,4,5,6} 
}

\begin{document}
\maketitle
\newcommand\imgdim{40mm}
\

{\footnotesize \textsuperscript{1}Institute of Medical Science, University of Toronto, Toronto, ON, Canada\\
\textsuperscript{2}Department of Diagnostic \& Interventional Radiology, Research Institute, The Hospital for Sick Children, Toronto, ON, Canada\\
\textsuperscript{3}Department of Medical Imaging, University of Toronto, Toronto, ON\\
\textsuperscript{4}Department of Computer Science, University of Toronto, Toronto, ON, Canada\\
\textsuperscript{5}Department of Mechanical and Industrial Engineering, University of Toronto\\
\textsuperscript{6}Vector Institute, Toronto, ON, Canada\\}

\begin{abstract}

Training machine learning models to segment tumors and other anomalies in medical images is an important step for developing diagnostic tools but generally requires manually annotated ground truth segmentations, which necessitates significant time and resources. This work proposes the use of a superpixel generation model and a superpixel clustering model to enable weakly supervised brain tumor segmentations. The proposed method utilizes binary image-level classification labels, which are readily accessible, to significantly improve the initial region of interest segmentations generated by standard weakly supervised methods without requiring ground truth annotations. We used 2D slices of magnetic resonance brain scans from the Multimodal Brain Tumor Segmentation Challenge 2020 dataset and labels indicating the presence of tumors to train the pipeline. On the test cohort, our method achieved a mean Dice coefficient of 0.691 and a mean 95\% Hausdorff distance of 18.1, outperforming existing superpixel-based weakly supervised segmentation methods. 


\end{abstract}

\section{Introduction}

Segmentation is crucial in medical imaging for localizing regions of interest (ROI), such as tumors, which can then assist in the identification of anomalies. Machine learning (ML) can automate the segmentation task with excellent performance, as demonstrated by top-performing models in the Multimodal Brain Tumor Segmentation Challenge (BraTS) 2020 challenge \citep{henry_brain_2020,10.1007/978-3-030-72087-2_11,10.1007/978-3-030-72087-2_6}. However, training ML segmentation models demands large datasets of manually annotated medical images which are not only tedious and expensive to acquire, but also may be inaccessible for specific diseases such as rare cancers. Weakly supervised training of segmentation models, which does not require segmentation labels, has great potential to localize anomalies only using image-level classification labels that are much less expensive to acquire than manual pixel-level annotations. 

Work into weakly supervised segmentation where the only available ground truths are image-level classification labels often involves training a classification model that is used to infer tumor segmentations through class activation maps. This approach has been used in a variety of medical imaging problems including the segmentation of organs \citep{Chen_2022_CVPR}, pulmonary nodules \citep{10.1007/978-3-319-66179-7_65}, and brain lesions \citep{10.1007/978-3-030-32248-9_24}.

Another weakly supervised segmentation approach is to utilize superpixels. Superpixels are pixels grouped based on various characteristics, including pixel gray levels and proximity. By grouping pixels together, superpixels capture redundancy and reduce the complexity of computer vision tasks making them valuable for image segmentation \citep{rs12061049,Kwak_Hong_Han_2017,YI2022108504}. A notable approach to ML-based superpixel segmentation uses a Fully Convolutional Network (FCN) to generate oversegmented superpixels with less computational complexity \citep{yang_superpixel_2020}. 

We hypothesize that superpixels can be leveraged to acquire additional contextual information to improve weakly supervised segmentations. We propose to simultaneously train a superpixel generation model and a superpixel clustering model using localization seeds acquired from a classifier trained with the image-level labels. For each pixel, the superpixel generator assigns association scores to each superpixel group, and the clustering model predicts weights for each superpixel based on their overlap with the tumor. Pixels are soft clustered based on their association with highly weighted superpixels to form segmentations. The superpixel models combine information from the pixel intensities with information from the localization seeds, yielding segmentations that are consistent with both the classifier understanding from the localization seeds and the pixel intensities of the MR images.

The novelty of the work is summarized by the following points:
\begin{itemize}
    \item Simultaneous deep superpixel generation and clustering enable effective weakly supervised segmentation of brain tumors on MRI datasets. 
    \item Localization seeds that undersegment the cancerous and non-cancerous regions are effective priors of information and can be generated from binary classifiers trained to identify cancerous images. 
    \item The use of deep superpixel generation and clustering improves segmentation performance and inference time over other superpixel-based methods. 
\end{itemize}

\section{Materials and methods}
\subsection{Dataset and Preprocessing}
To form our dataset, the 369 T1-weighted, post-contrast T1-weighted, T2-weighted, and T2 Fluid Attenuated Inversion Recovery (T2-FLAIR) MRI volumes from the BraTS 2020 dataset \citep{bakas_segmentation_2017-1,bakas_segmentation_2017,bakas_advancing_2017,Bakas_Reyes_Jakab_Bauer_Rempfler_Crimi_Shinohara_Berger_Ha_Rozycki_2019,menze_multimodal_2015} were stacked together. The stacked MRI volumes were then split into axial slices to form stacked 2-dimensional (2D) images with 4 channels. Only the training set of the BraTS dataset was used because it is the only one with publicly available ground truths. 

The images were preprocessed by first cropping each image and segmentation map using the smallest bounding box which contained the brain, clipping all non-zero intensity values to their 1 and 99 percentiles to remove outliers, normalizing the cropped images using min-max scaling, and then randomly cropping the images to fixed patches of size $128 \times 128$ along the coronal and sagittal axes, as done by Henry et al. \citep{henry_brain_2020} and Wang et al. \citep{wang_3d_2020} in their work with BraTS datasets. The 369 available patient volumes were then split into 295 (80\%), 37 (10\%), and 37 (10\%) volumes for the training, validation, and test cohorts, respectively. After splitting the volumes into 2D images, the first 30 and last 30 slices of each volume were removed, as done by Han et al. \citep{8869751} because these slices lack useful information. The training, validation, and test cohorts had 24635, 3095, and 3077 stacked 2D images, respectively. For the training, validation, and test cohorts, respectively; 68.9\%, 66.3\%, and 72.3\% of images were cancerous. The images will be referred to as $x_k \in \mathbb{R}^{4, H, W}$, where $H=128$ and $W=128$. Ground truths for each slice $y_k$ were assigned 0 if the segmentations were empty, and $1$ otherwise. 

\subsection{Proposed Weakly Supervised Segmentation Method}

\begin{figure}[h!]
\begin{center}
\includegraphics[width=\textwidth]{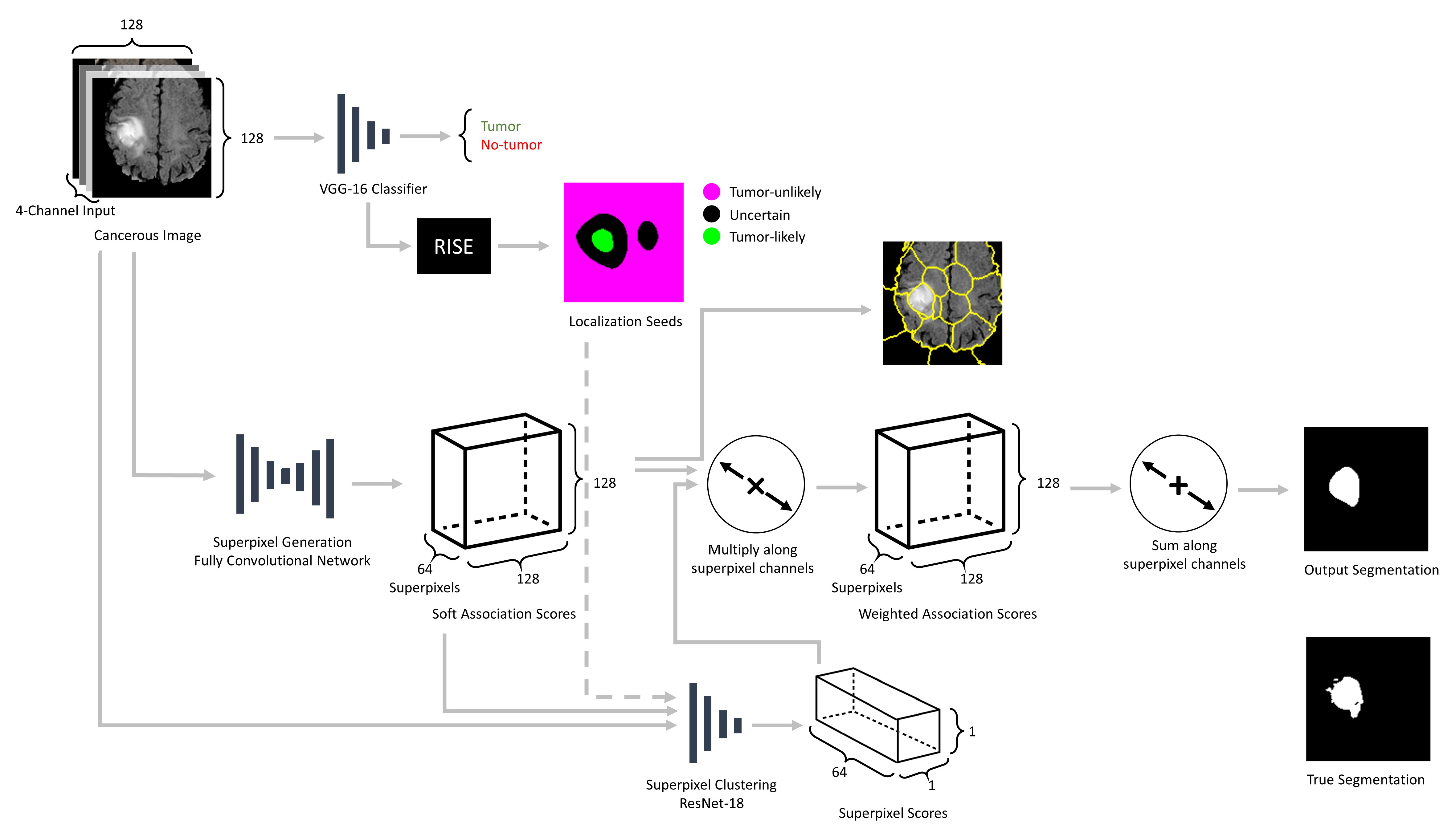}
\end{center}
\caption{Flowchart of proposed weakly supervised segmentation method. For the localization seeds component; green indicates positive seeds, magenta indicates negative seeds, black indicates unseeded regions. Solid lines represent use as inputs and outputs. Dashed lines represent use in loss functions.} \label{refinement_flowchart}
\end{figure}

We first trained a classifier model to identify whether an image contains a tumor, then generated localization seeds from the model using Randomized Input Sampling for Explanation of Black-box Models (RISE) \citep{Petsiuk2018RISERI}. The localization seeds use the classifier's understanding to split each pixel of the images into one of three categories. The first, referred to as positive seeds, indicates regions of the image with high likelihood of containing a tumor. The second, referred to as negative seeds, indicates regions with low likelihood of containing a tumor. The final category, referred to as unseeded regions, corresponds to the remaining areas of the images and indicates regions of low confidence from the classifier. This results in positive seeds that undersegment the tumor, and negative seeds that undersegment the non-cancerous regions. Assuming that the positive and negative seeds are accurate, these seeds simplify the task to classifying the unseeded regions as overlapping with or not overlapping with a tumor. The seeds are used as pseudo-ground truths to simultaneously train both a superpixel generator and a superpixel clustering model which, when used together, can produce the final refined segmentations from the probability heat map of the superpixel-based segmentations. A flowchart of the proposed methodology is presented in Figure \ref{refinement_flowchart}. We chose to use 2D images over 3D images because converting 3D MRI volumes to 2D MR images yields significantly more data samples and reduces memory costs. Furthermore, previous work demonstrated that brain tumors can be effectively segmented from 2D images \citep{8964956}.

\subsubsection{Classifier Model}
The classifier model uses a VGG-16 architecture \citep{simonyan2014very} with batch normalization, whose output is passed through a Sigmoid function to generate the probability that each $x_k \in X$ contains a tumor, where $X = \{x_1, x_2, ..., x_N\} \in \mathbb{R}^{N, H, W}$ is a set of brain MR images. Prior to being input to the classifier model, the images are upsampled by a factor of 2. The images are not upsampled for any other model in the proposed method. This classifier model is trained using $Y = \{y_1, ..., y_N\}$ as the ground truths, where $y_k$ is a binary label with a value of 1 if $x_k$ contains tumor and 0 otherwise. The remainder of the method only uses images predicted by the classifier to contain tumors, to avoid attempting to segment healthy images. This subset of $X$ will be referred to as $\hat{X} = \{\hat{x}_1, ..., \hat{x}_{\hat{N}} \}$. The methodology is independent of the VGG-16 architecture, and thus, other classifier architectures can be used instead. 

The classifier was trained to optimize the binary cross-entropy between the output probabilities and the binary ground truths using Adam optimizer with $\beta_1 = 0.9, \beta_2 = 0.999, \epsilon=1e-8$, and a weight decay of $0.1$ \citep{Kingma_Ba_2017}. The classifier was trained for 100 epochs using a batch size of 32. The learning rate was initially set to $5e-4$ and then decreased by a factor of 10 when the validation loss did not decrease by $1e-4$.

\subsubsection{RISE Method} 
RISE \citep{Petsiuk2018RISERI} is used to generate heat maps $H_{rise} \in \mathbb{R}^{\hat{N}, H, W}$ for each of the images predicted to be cancerous. The heat maps indicate the approximate likelihood for tumors to be present at each pixel. These heat maps were converted to localization seeds by setting the pixels corresponding to the top 20\% of values in $H_{rise}$ as positive seeds, and setting the pixels corresponding to the bottom 20\% of values as negative seeds. $S_+ = \{s_{+_1}, s_{+_2}, ..., s_{+_{\hat{N}}}\} \in \mathbb{R}^{\hat{N}, H, W}$ is a binary map indicating positive seeds and $S_- = \{s_{-_1}, s_{-_2}, ..., s_{-_{\hat{N}}}\} \in \mathbb{R}^{\hat{N}, H, W}$ is a binary map indicating negative seeds. When using RISE, we set the number of masks for an image to 4000 and use the same masks across all images.

\subsubsection{Proposed Superpixel Generation and Clustering Models}
The superpixel generation model and the superpixel clustering model are used to output the final segmentations without using the ground truth segmentations. The superpixel generation model assigns $N_S$ soft association scores to each pixel, where $N_S$ is the maximum number of superpixels to generate, which we set to 64. The association maps are represented by $Q = \{q_1, ..., q_{\hat{N}}\} \in \mathbb{R}^{\hat{N}, N_S, H, W}$, where $q_{k, s, p_y, p_x}$ is the probability that the pixel at $(p_y, p_x)$ is assigned to the superpixel $s$. Soft associations may result in a pixel having similar associations to multiple superpixels. The superpixel clustering model then assigns superpixel scores to each superpixel indicating the likelihood that each superpixel represents a cancerous region. The superpixel scores are represented by $R = \{r_1, ..., r_{\hat{N}}\} \in \mathbb{R}^{\hat{N}, N_S}$ where $r_{k, s}$ represents the probability that superpixel $s$ contains a tumor. The pixels can then be soft clustered into a tumor segmentation by performing a weighted sum along the superpixel association scores using the superpixel scores as weights. The result of the weighted sum is the likelihood that each pixel belongs to a tumor segmentation based on its association with strongly weighted superpixels.

The superpixel generator takes input $\hat{x}_k$ and outputs a corresponding value $q_k$ by passing the direct output of the superpixel generation model through a SoftMax function to rescale the outputs from 0 to 1 along the $N_s$ superpixel associations. The clustering model uses a ResNet-18 architecture \citep{he_deep_2015} and receives a concatenation of $\hat{x}_k$ and $q_k$ as input. The outputs of the clustering model are passed through a SoftMax function to yield superpixel scores $R$. Heatmaps $H_{spixel_+} \in \mathbb{R}^{\hat{N}, H, W}$ that localize the tumors can be acquired from $Q$ and $R$ by multiplying each of the $N_S$ association maps in $Q$ by their corresponding scores $R$, and then summing along the $N_S$ channels as shown in Equation \ref{H_spixel_pos_eq}. The superpixel generator architecture is based on AINet proposed by Wang et al. \citep{wang_ainet:_2021}, which is a Fully Convolutional Network (FCN)-based superpixel segmentation model that uses a variational autoencoder (VAE). Unlike AINet, which outputs local superpixel associations, we use global associations so that $Q$ can be passed into the superpixel clustering model. This allows the generator model to be trained in tandem with the clustering model. The training uses two different loss functions. The first loss function, $L_{spixel_+}$, was proposed by Yang et al. \citep{yang_superpixel_2020} and minimizes the variation in pixel intensities and pixel positions in each superpixel. This loss is defined in Eq.~\ref{superpixel_loss}, where $p$ represents a pixel's coordinates ranging from $(1, 1)$ to $(H, W)$, and $m$ is a coefficient used to tune the size of the superpixels, which we set as $\frac{3}{160}$. We selected this value for $m$ by multiplying the value suggested by the original work, $\frac{3}{16000}$ \citep{yang_superpixel_2020}, by 100 to achieve the desired superpixel size. $l_s$ and $u_s$ are the vectors representing the mean superpixel location and the mean superpixel intensity for superpixel $s$, respectively. The second loss function, $L_{seed}$, is a loss from the Seed, Expand, and Constrain paradigm for weakly supervised segmentation. This loss is designed to train models to output segmentations that include positive seeded regions and exclude negative seeded regions \citep{kolesnikov2016seed}. This loss is defined in Eq.~\ref{H_spixel_pos_eq}-\ref{seed_eq} where C indicates whether the positive or negative seeds of an image $s_k$ is being evaluated. These losses, when combined together, encourage the models to account for both the localization seeds $S$ and the pixel intensities. This results in $H_{spixel_+}$ localizing the unseeded regions that correspond to the pixel intensities in the positive seeds. The combined loss is presented in Eq.~\ref{combined_loss_eq}, where $\alpha$ is a weight for the seed loss. The output $H_{spixel_+}$ can then be thresholded to generate final segmentations $E_{spixel_+} \in \mathbb{R}^{\hat{N}, H, W}$.

The superpixel generation and clustering models were trained using an Adam optimizer with $\beta_1 = 0.9, \beta_2 = 0.999, \epsilon=1e-8$, a weight decay of $0.1$. The models were trained for 100 epochs using a batch size of 32. The learning rate was initially set to $5e-4$, which was halved every 25 epochs. The weight for the seed loss, $\alpha$, was set to $50$.

\begin{equation}
\label{H_spixel_pos_eq}
H_{{spixel_+}_k} = \sum_{s \in N_s} Q_{k,s} R_{k,s}
\end{equation}

\begin{equation}
\label{superpixel_loss}
L_{spixel} = \frac{1}{\hat{N}} \sum_{k=1}^{\hat{N}} \sum_p \left( \left\Vert \sum_{s \in N_s} u_s Q_{k,s}(p) \right\Vert_2 + m \left\Vert \sum_{s \in N_s} l_s Q_{k,s}(p) \right\Vert_2 \right)
\end{equation}

\begin{equation}
\label{H_spixel_neg_eq}
H_{{spixel_-}_k} = 1 - H_{{spixel_+}_k}
\end{equation}

\begin{equation}
\label{seed_eq}
L_{seed} = \frac{1}{\hat{N}} \sum_{k=0}^{\hat{N} - 1} \left( \frac{-1}{\sum_{C \in [+, -]} \lvert {s_{C_k}} \rvert} \sum_{C \in [+, -]} \sum_{i, j \in s_{C_k}} \log \left( {H_{{spixel_C}_k}}_{i,j} \right) \right)
\end{equation}

\begin{equation}
\label{combined_loss_eq}
L = L_{spixel} + \alpha L_{seed}
\end{equation}

\section{Results}
\label{sec:results}

We trained our models using images $X$ and binary image-level labels $Y$ without using any segmentation ground truths. The classifier achieved training, validation, and test accuracies of $0.996$, $0.912$, and $0.933$, respectively, using a decision threshold of $0.5$. Table \ref{dice_coefficients} presents the per-image mean Dice coefficients (Dice) and the 95\% Hausdorff distance (HD95) between the output segmentations for our proposed method and the ground truth segmentations. 

\begin{table}[h!]
\begin{center}
\caption{Dice coefficients and 95\% Hausdorff distances between generated segmentations and true segmentations.}\label{dice_coefficients}
\begin{tabular}{lcccccccc}
\toprule
\multirow{2}{*}[-2pt]{\makecell{Method}} & \multicolumn{4}{c}{Dice} & \multicolumn{4}{c}{HD95} \\
\cmidrule(lr){2-5} \cmidrule(lr){6-9}
& \multirow{2}{*}[-2pt]{\makecell{Training}} & \multirow{2}{*}[-2pt]{\makecell{Validation}} & \multirow{2}{*}[-2pt]{\makecell{Test}} & \multirow{2}{*}[-2pt]{\makecell{BraTS\\2023}} & \multirow{2}{*}[-2pt]{\makecell{Training}} & \multirow{2}{*}[-2pt]{\makecell{Validation}} & \multirow{2}{*}[-2pt]{\makecell{Test}} & \multirow{2}{*}[-2pt]{\makecell{BraTS\\2023}} \\
\\
\cmidrule(lr){1-1} \cmidrule(lr){2-2} \cmidrule(lr){3-3} \cmidrule(lr){4-4} \cmidrule(lr){5-5} \cmidrule(lr){6-6} \cmidrule(lr){7-7} \cmidrule(lr){8-8} \cmidrule(lr){9-9}
Proposed  ($\alpha$ = 50) & \textbf{0.733} & \textbf{0.715} & \textbf{0.691} & \textbf{0.745} & \textbf{16.5} & \textbf{13.4} & \textbf{18.1} & \textbf{20.8} \\ 
Proposed ($\alpha$ = 10) & 0.609 & 0.608 & 0.594 & 0.574 & 20.6 & 17.9 & 24.9 & 34.7 \\
Ablation & 0.710 & 0.697 & 0.671 & 0.697 & 18.6 & 13.7 & 18.6 & 22.2 \\
PatchConvNet & 0.185 & 0.159 & 0.134 & 0.001 & 53.9 & 49.7 & 54.13 & 87.3 \\
\midrule
SPN & 0.125 & 0.117 & 0.139 & 0.262 & 57.6 & 53.2 & 58.0 & 74.2 \\
SPN (classifier) & 0.423 & 0.394 & 0.375 & 0.260 & 55.7 & 48.6 & 53.9 & 73.4 \\
MIL & 0.190 & 0.209 & 0.199 & 0.108 & 25.7 & 24.2 & 25.5 & 49.6 \\
MIL (classifier) & 0.426 & 0.403 & 0.391 & 0.126 & 47.3 & 40.3 & 47.6 & 53.4 \\
\bottomrule
\end{tabular}
\end{center}
\end{table}

We also present the performance of baseline methods for comparison. The first baseline method is the proposed method using a seed loss weight of $10$ rather than a seed loss weight of $50$. This is to determine the impact of the seed loss weight on the segmentation performance. The second baseline method is the performance of the AINet architecture used by the superpixel generator model with the superpixel components removed and altered to directly output segmentations. This method, referred to as ablation, serves as an ablation study that investigates the impact of the superpixel component of the proposed method. The third baseline method is our proposed method with the VGG-16 classifier replaced by a PatchConvNet classifier \citep{touvron2021patchconvnet}. PatchConvNet is a more recent classifier that is designed to generate accurate attention maps, which we used as the seeds in place of the RISE generated seeds for this baseline method.

In addition, we also compared our proposed method to two other methods designed for weakly supervised segmentation. The first is the Superpixel Pooling Network (SPN), proposed by Kwak et al., which uses pre-generated superpixels to perform weakly supervised segmentation \citep{Kwak_Hong_Han_2017}. This method relies on pre-generated superpixels, which we generated using Felzenszwalb's Algorithm using a scale of 100 and a standard deviation of 0.8. We chose these hyperparameters as they set the number of output superpixels to approximately 100, thereby decreasing training time. The second is a Multiple Instance Learning (MIL) method proposed by Lerousseau et al. \citep{lerousseau:hal-03133239}. MIL involves training a learning model using instances arranged in sets, patches of an image in this case, and then aggregating the predictions of the instances to output a prediction for the whole set. To train the MIL baseline, we used a VGG-16 model with batch normalization. At each epoch, we extracted 50 patches of shape $128 \times 128$ from the images after upsampling them to $512 \times 512$. At each iteration, we set the 20\% of patches with the highest predictions to be cancerous and 20\% of patches with the lowest predictions to be non-cancerous, as these thresholds were demonstrated to be effective in the original work and are consistent with the thresholds we used when generating seeds using RISE.

The SPN and MIL methods differ from the other baseline methods in that they are not variants of the proposed method, and thus do not assign empty segmentations to images classified as non-cancerous. To allow for effective comparison, we present the performance of these two baseline methods with and without using a classifier to assign empty segmentations. The results in Table \ref{dice_coefficients} for SPN and MIL using the classifier are noted by the term (classifier). For these results, we used the VGG-16 classifier trained for our proposed method.

We assess the generalizability of the proposed method by evaluating each trained model on the BraTS 2023 dataset \citep{baid2021rsnaasnrmiccai,menze_multimodal_2015,bakas_advancing_2017,bakas_segmentation_2017-1,bakas_segmentation_2017}. To do so, we removed data that appeared in the BraTS 2020, preprocessed the images as detailed in the methodology, and then extracted the cross-section with the largest tumor area from each patient. This resulted in 886 images from the BraTS 2023 dataset. The performance of each model on these images can be found under the BraTS 2023 columns in Table \ref{dice_coefficients}.

Each of the presented methods uses a decision threshold to convert the output probability maps to binary segmentations. The decision threshold for each method was determined by evaluating the Dice on the validation cohort at threshold intervals of 0.1 and choosing the threshold that yielded the maximum validation Dice. The proposed methods used thresholds of 0.6 and 0.9 for seed loss weights of 50 and 10, respectively, while the ablation and PatchConvNet methods used thresholds of 0.5 and 0.9, respectively. Both SPN models used threshold of 0.9 while the MIL models used a threshold of 0.3 when using the classifier and a threshold of 0.9 when not using the classifier.

\begin{figure}
\begin{center}
\includegraphics[width=0.9\textwidth]{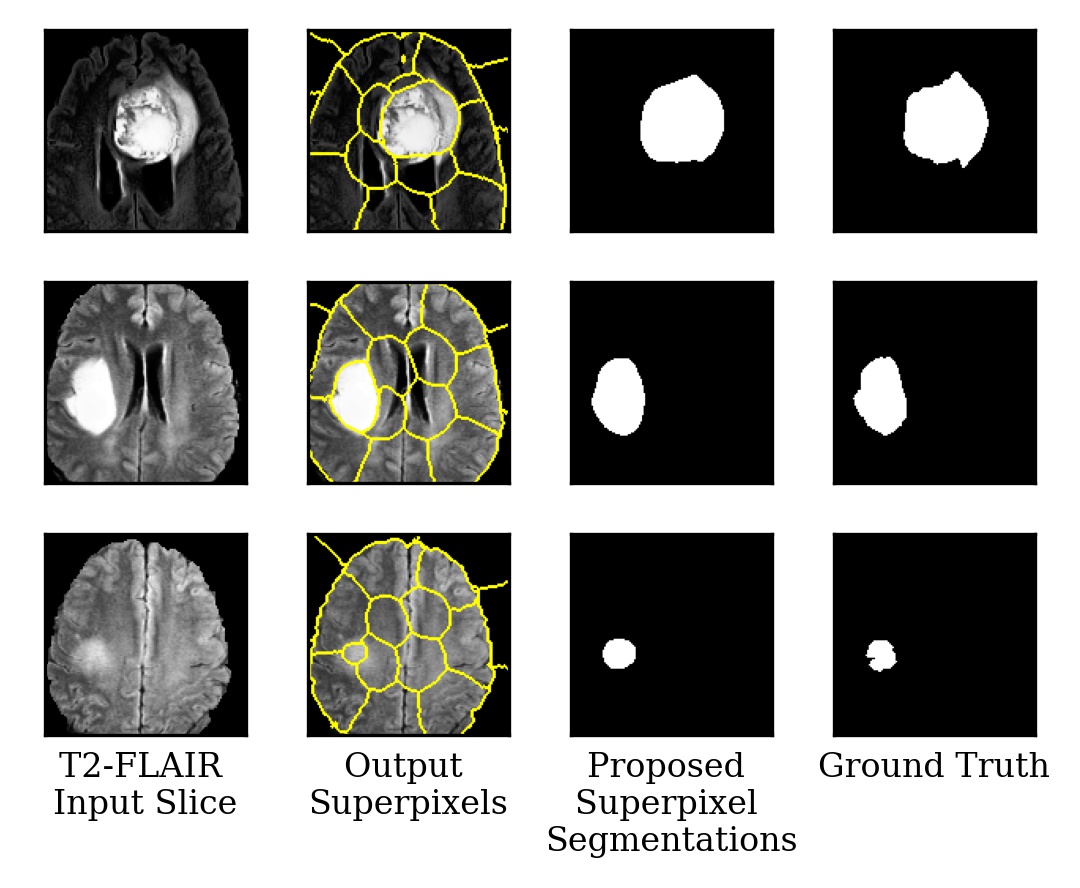}
\caption{Visualization of T2-FLAIR channel of MR images, generated superpixels, output segmentations, and true segmentations for three examples.}
\label{segmentation_refinements}
\end{center}
\end{figure}

Figure \ref{segmentation_refinements} presents three images from the test set and their corresponding segmentations generated at each step of the pipeline, as well as the ground truth segmentations.

When evaluating the inference time of the proposed method compared to SPN, the proposed method had an average inference time of 5.85 milliseconds while the SPN method had an average inference time of 28.5 milliseconds and the MIL method had an average inference time of 1.77 milliseconds per patch.

\section{Discussion}

\subsection{Key Findings}

When comparing the performance of the proposed method to the SPN and MIL baseline methods, the proposed method and the ablation method outperformed SPN and MIL in both Dice and HD95. The improved performance indicates that the SPN and MIL methods, while being effective in tasks with large training datasets, can worsen in tasks with limited available data such as brain tumor segmentation. MIL is frequently used for weakly supervised segmentation of histopathology images because of the need to interpret the large gigapixel resolution images in patches. We believe the significantly reduced spatial dimensions and resolutions of the MR images negatively impacted the performance of the MIL baseline. The MR images lacked the resolution required to extract patches with sufficient information that only occupied a small portion of its source image. As such, the MIL baseline was unable to effectively learn to segment the tumors.

PatchConvNet also suffered from the smaller dataset size. The PatchConvNet classifier was not able to generate effective undersegmented positive and negative seeds to guide the training of the superpixel generator and clustering models. This can be attributed to the smaller dataset size, which PatchConvNet was not designed for, and the use of attention-based maps for the seeds. With the smaller dataset size, PatchConvNet was unable to acquire an effective understanding of the tumors. As a result, the attention maps acquired from PatchConvNet did not consistently undersegment the cancerous and non-cancerous regions, which is a critical assumption when using the seeds. Using a VGG-16 classifier and generating the seeds using RISE resulted in localization seeds that tend to undersegment the cancerous and non-cancerous regions despite the limited available data. In contexts with more available training data, PatchConvNet could be used to generate effective seeds but PatchConvNet seems to struggle in tasks with small dataset sizes, which are very common in medical contexts.

Superpixels generated from algorithmic methods such as 
Simple Linear Iterative Clustering or Felzenszwalb segmentation have previously been used for weakly supervised segmentation, often in conjunction with CAMs, \citep{rs12061049,Kwak_Hong_Han_2017,YI2022108504}. The SPN is one such approach. Unlike the proposed method, SPN uses pre-generated superpixels and produces the CAMs by outputting weights for each superpixel and grouping the the highest weighted superpixels together. 

The use of simultaneously generated superpixels is a key novelty of our work. When using traditional superpixel generation algorithms, the precision of the segmentations is dependent on the number of superpixels, as fewer superpixels can result in less refined boundaries. However, such superpixel generation algorithms lack a means of setting a consistent number of superpixels across all images. Accounting for the varying number of superpixels leads to significantly increased computational complexity. This is demonstrated by how the inference time of our proposed method was 79.47\% faster than the inference time of the SPN. 

Using a consistent number of superpixels across images for pre-generated superpixels can raise concerns regarding the quality of the segmentation boundaries. Training a deep learning model to generate superpixels simultaneously with a superpixel clustering model allows for the gradients of the loss functions encouraging accurate segmentations to propagate through the superpixel generation model as well. This helps the superpixel generation model to not just learn to generate superpixels, but to generate superpixels with refined boundaries around the tumors. Thus, simultaneous generation and clustering of superpixels using neural networks improves the inference time and the segmentation performance when using superpixels for segmentation. 

Figure \ref{segmentation_refinements} demonstrates how the proposed method can reduce the number of outputted superpixels despite using a predefined number of superpixels. In our test cohort, the models reduced the number of superpixels from a predefined limit of 64 to approximately 22 per image by outputting 64 superpixels but having the majority of superpixels have no associated pixels.

In Figure \ref{segmentation_refinements}, the superpixels do not perfectly contour the segmented regions because the segmentations are calculated using a weighted sum of the superpixel scores based on their association with each pixel. Thus, pixels whose most associated superpixel is not primarily a part of the segmented region can be segmented so long as it has a sufficiently high association score with the primarily segmented superpixel. As such, the segmentations cannot be generated simply by selecting superpixels outputted by the method, they need to be soft clustered using the superpixel association and weights. Despite the lower number of superpixels when using higher seed loss weights, the method is still able to segment smaller tumors. It can also be seen that superpixels outside of the tumor regions do not align with brain subregions or local patterns. This indicates that the superpixels are tuned to segment specifically brain tumors. While Figure \ref{segmentation_refinements} implies that only one superpixel is approximately required for each image, we argue that the clustering component has the benefit of allowing this method to be applied to tasks with multiple localized anomalies in each image.

\subsection{Limitations}

A limitation of this method is its reliance on superpixels which are computed based on pixel intensity. While the superpixels provide valuable information that improve segmentations of brain tumors, the superpixels also provide constraints on the set of problems this method can be applied to. In particular, this method would be ineffective for segmenting non-focal ROIs.

In addition, the proposed method relies on the localization seeds to be trained effectively. Despite not requiring the localization seeds during inference, poor localization seeds during training can propagate errors leading to poor segmentations during inference. The performance of the PatchConvNet baseline demonstrates the importance of seed accuracy. PatchConvNet was unable to output effective localization seeds and using the seeds from PatchConvNet with our proposed method decreased the Dice coefficient from 0.691 to 0.134 on the test cohort. As such, effective localization seeds from an accurate classifier that undersegment the cancerous and non-cancerous regions are crucial for effective performance using the proposed method.

Another limitation is that this method cannot be trained end-to-end. While the method is a weakly supervised approach as it does not require any segmentation ground truths to train, it can also be interpreted as a fully supervised classification task followed by an unsupervised superpixel generation and clustering task. Without having seeds generated from an accurate classifier to guide the downstream models, crucial information that informs the segmentation task is lost. Many clinical contexts have classifiers already available that can be applied to this method. However, the proposed method cannot be applied to contexts without readily available classifiers that require end-to-end training.

A shortcoming of this study is its use of 2D images rather than 3D images due to the GPU memory costs required to generate 3D superpixels using an FCN-based superpixel generation model. The method is not limited to 2D images and thus it is of interest to explore applications of this method in 3D contexts. Previous studies have demonstrated that 3D segmentation leads to superior performance compared to 2D segmentation, which suggests that this method could be improved further when applied to 3D images \citep{Avesta_Hossain_Lin_Aboian_Krumholz_Aneja_2023}. 

As is the case with other weakly supervised segmentation methods, the performance of our proposed method does not match the performance of fully supervised methods. 3D Fully supervised methods have achieved Dice coefficients ranging from 0.88 to 0.92 on the test cohort of the BraTS 2020 dataset \citep{10.1007/978-3-030-72087-2_11, jia2020h2nfnet, wang2020modalitypairing, yuan2020automatic}. However, until weakly supervised segmentations match the performance of fully supervised approaches, weakly supervised segmentation methods serve a different purpose than fully supervised segmentation methods. Weakly supervised segmentations are very effective at generating initial segmentations that can be revised by radiologists or for downstream semi-supervised training to reduce workload on medical datasets that lack manual annotations. In summary, despite the lower performance of our proposed method compared to fully supervised methods, our proposed method is effective for generating initial segmentations when manually annotated training data is not available.

\subsection{Conclusion}
We introduced a weakly supervised superpixel-based approach to segmentation that incorporates contextual information through simultaneous superpixel generation and clustering. Integrating superpixels with localization seeds provides information on the boundaries of the tumors, allowing for the segmentation of tumors only using image-level labels. We demonstrated that generating superpixels using a deep learning model during training is not only faster but also yields improved segmentations compared to using superpixels generated from traditional approaches. This work can be used to improve the development of future weakly supervised segmentation methods through the integration of superpixels.

\section*{Conflict of Interest Statement}

The authors declare that the research was conducted in the absence of any commercial or financial relationships that could be construed as a potential conflict of interest.

\section*{Author Contributions}


JY, KN, and FK contributed to the design of the concept, study, and analysis. JY implemented the algorithms, conducted the experiments, and wrote the original draft. All authors contributed to the writing and editing of the manuscript and approved the final manuscript. FK acquired the funding, and supervised the research.

\section*{Funding}
This study received funding from Huawei Technologies Canada Co., Ltd. The funder was not involved in the study design, collection, analysis, interpretation of data, the writing of this article or the decision to submit it for publication. All authors declare no other competing interests.

\bibliography{refs}


\end{document}